\documentclass{ifacconf}

\usepackage{graphicx}      
\usepackage{natbib}        
\usepackage{mathtools}
\usepackage{amsmath,amssymb,amsfonts}
\usepackage{algorithmic}
\usepackage{graphicx}
\usepackage{textcomp}
\usepackage[OT1]{fontenc}
\usepackage{pifont}
\usepackage{caption}  
\usepackage{subcaption}
\usepackage{multirow}
\usepackage{scalerel}
\usepackage{threeparttable}
\usepackage{booktabs}
\usepackage[dvipsnames,svgnames]{xcolor, colortbl}
\begin{document}
\begin{frontmatter}

\title{Robust Control using Control Lyapunov Function and Hamilton-Jacobi  Reachability
} 
%
\thanks[footnoteinfo]{ The work was supported by NSF grant 2128568}
\vspace{-0.75cm}
\author{Chun-Ming Yang, Pranav A. Bhounsule}
\vspace{-0.25cm}

\address{Department of Mechanical and Industrial Engineering, \\ University of Illinois at Chicago, 
       842 W Taylor St, Chicago, IL 60607, USA. (e-mail: jyang241@uic.edu; pranav@uic.edu)
       }
       
\begin{abstract}                
The paper presents a robust control technique that combines the Control Lyapunov function and Hamilton-Jacobi Reachability to compute a controller and its Region of Attraction (ROA). The Control Lyapunov function uses a linear system model with an assumed additive uncertainty to calculate a control gain and the level sets of the ROA as a function of the uncertainty. Next, Hamilton-Jacobi reachability uses the nonlinear model with the modeled uncertainty, which need not be additive, to compute the backward reachable set (BRS). Finally, by juxtaposing the level sets of the ROA with BRS, we can calculate the worst-case additive disturbance and the ROA of the nonlinear model. We illustrate our approach on a 2D quadcopter tracking trajectory and a 2D quadcopter with height and velocity regulation in simulation.
\end{abstract}


\end{frontmatter}

\section{Introduction}
Model predictive control (MPC), an online optimal control technique, has received increased adaptation because of its ability to quickly adapt to the situation in real-time  \citep{di2018dynamic,7505151}. However, MPC may perform poorly when there is model uncertainty. Hence there is a need to investigate techniques to enable robustification of MPC controllers. 

Robust control offers the best possible control for the worst-case uncertainty. The H infinity method is a robust control approach for linear or linearized systems but require a fairly accurate model \citep{babar2013robust}. On the other hand, adaptive control uses experimental data to either update the model or the controller \citep{sombolestan2023hierarchical}. Learning approaches such as Reinforcement learning can provide robustness using  dynamic randomization. Here the controller is a neural network that is tuned by varying the model and environmental parameters in the simulator, thus making the controller robust to uncertainty \citep{tan2018sim}. While robust and adaptive control offers stability guarantees, learning methods provide no such guarantees.

The safety of controllers is often guaranteed by computing the region of attraction (ROA). The sum-of-squares method uses a polynomial approximation of the system dynamics and convex optimization to estimate the ROA \citep{tedrake2009lqr}. The control-Lyapunov and control-barrier function (CLF-CBF) approaches do not compute the ROA. Still, they guarantee that the system would not violate safety constraints through the barrier function and ensure reachability to the goal using the Lyapunov function \citep{ames2016control}. More recently CLF-CBF has been combined with adaptive control to enable robust control under uncertainty \citep{minniti2021adaptive}. The Hamilton-Jacobi (HJ) reachability analysis is a tool to compute the target sets the system can reach from a given state under constraints on the control and disturbances. It has been used to calculate the safety regions for quadcopter \citep{aswani2013provably} and path planning of vehicles \citep{chen2018robust}.

\begin{figure}
\centering
\centering\includegraphics[scale=1]{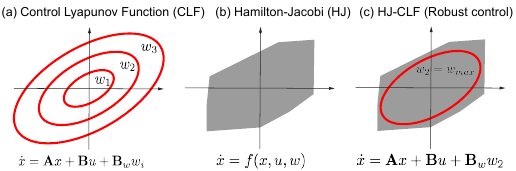}
\caption{(a) Control Lyapunov function (CLF) uses the linear dynamics to compute the controller and the level set for the Region of Attraction (ROA) for a given maximum disturbance $w_i$ (red ellipses); (b) Hamilton-Jacobi (HJ) Reachability uses the nonlinear dynamics with the disturbance $w$ to compute the Backward Reachable set (gray region); (c) Superimposing the BRS with level sets enables computation of the worst case disturbance (here $w_2=w_{max}$) and the safe set (red ellipse)}
\label{fig:overview}
\end{figure}

In this paper, we introduce Robust Model Predictive Control as a method to do  planning with safety guarantees for bounded model uncertainty. We assume that there is a nominal MPC controller that works for the uncertainty-free nominal system. We linearize the system, and use a Control Lyapunov Function approach to compute the feedback control gain and the level sets for the ROA for different values of the additive disturbances (see Fig.~\ref{fig:overview} (a)). Note that the additive disturbance in the linearized model is an assumption as the actual model may not have an additive disturbance. We use the nonlinear model with the Hamilton Jacobi (HJ) reachability to compute the safe set (see Fig.~\ref{fig:overview} (b)). Finally, by super-imposing the level sets on the safe set, we can compute the worst case additive disturbance for the linear model as well as the ROA (see Fig.~\ref{fig:overview} (c)). Our CLF method is similar to \citep{yu2010robust}, but the novelty lies in using HJ to estimate the RoA. The technique is demonstrated on a 2D quadcopter tracking a trajectory under disturbance and a 2D quadruped with regulating the height in the presence of added weight and tracking a horizontal speed while pushing a box.


\section{Methods} \label{sec:methods}

\subsection{Model Predictive Control (MPC)} \label{sec:MPC}
Model predictive control (MPC) is an online optimization technique. It uses the current estimate of the state and a model of the plant to estimate the state over a time horizon and a projected cost. The minimization of the cost gives the control over the time horizon. 
However, only the first computed control is applied, and the process is repeated for the next time step and so on.  

The MPC problem at time $t$ and with horizon length \(k\) can be written as follows
\begin{align} \label{eq:mpc_cost}
&\min_{x,u}\sum_{i=0}^{k-1}\left\| x(t+i)-x_{ref}(t) \right\|_{Q_{MPC}} + \left\| u(t+i)) \right\|_{R_{MPC}} \\
 \text{s.t.} &\quad \dot{x}_{t+i} = f(x_{t+i}, u_{t+i}), \\ 
 & \quad  x_{t+i+1}=x_{t+i} +  f(x_{t+i}, u_{t+i})\Delta t_{\scaleto{MPC}{3pt}}, \\
 & \quad  u_{i} \in \mathbb{U}, x_{t+i} \in \mathbb{X}, i=0,1...k-1
\end{align}
where \(x \in \mathbb{R}^{n}\) is the system state, \(u \in \mathbb{R}^{m}\) is the control input, \(Q_{MPC} \in \mathbb{R}^{n \times n} \) and \(R_{MPC} \in  \mathbb{R}^{m \times m} \) are user-chosen diagonal positive weight matrix, scalar \(\Delta t_{\scaleto{MPC}{3pt}}\) is the time step, and \( \mathbb{U}, \mathbb{X}\) are feasible sets for the control and state.

\subsection{Hamilton-Jacobi Reachability Analysis (HJ)} \label{sec:HJ}
Hamilton-Jacobi Reachability Analysis (HJ) provides formal guarantees for the performance and safety properties of nonlinear control systems. Here HJ is used to compute the backward reachable set (BRS). BRS \(\mathcal{R}(t)\) is the set of states \(x \in \mathbb{R}^{n}\) from which the system can be driven into the target set of states \(\mathcal{T}\) in a predefined time. 

Consider the dynamical system
\begin{equation} \label{eq:actual_dynamic}
\dot{x}(t) = f(x(t), u(t), w(t)) 
\end{equation}
with state $ x \in \mathbb{R}^n$, time $t$, control $u$, disturbance $w$, and dynamics $f: \mathbb{R}^n \times \mathcal{U} \times \mathcal{W} \rightarrow \mathbb{R}^n$ is assumed to be continuous, bounded, and Lipschitz continuous with respect to all inputs.
The disturbance is bounded as given below:
\begin{equation} \label{eq:disturbance_bound}
\mathbb{W} = \{w(t) \in \mathbb{R}^p \mid \|w(t)\|_{\infty} \leq w_{\text{max}}\}
\end{equation}

Let the solution to the Eqn.~\ref{eq:actual_dynamic} be $\chi(t;x,t_0,u(\cdot),w(\cdot)) : [t_0,0] \rightarrow \mathbb{R}^n$. The initial condition is $\chi(t_0;x,t_0,u(\cdot),w(\cdot)) = x_0$.

We define the target set of states as $\mathcal{T}$. The BRS is the set of states $\mathcal{R}(t)$ such that we can compute a control $u$ that will drive the $\mathcal{R}(t)$ to the $\mathcal{T}$ for the worst case disturbance $w_{max}$ exactly in time $t_0$. The BRS is defined mathematically as follows
\begin{align} 
\mathcal{R}(t) =\{ x:\forall u(\cdot) \in \mathbb{U}, \exists w(\cdot) \in \mathbb{W}, \exists t \in [t_0,0], ... \nonumber \\ 
 \chi(t;x,t_0,u(\cdot),w(\cdot))\in \mathcal{T} \} \label{eq:BRS}
\end{align}

In the HJ formulation, the target set is a sublevel set of a function $l(x)$, where $x\in\mathcal{T} \Leftrightarrow l(x) \leq 0$. The BRS in HJ reachability becomes the sublevel set of the value function $V(x,t)$ defined as follows
\begin{align}
V(x,t) := \underset{w(\cdot)}{\mbox{min}} \hspace{0.1cm} \underset{u(\cdot)} {\mbox{max}} \hspace{0.1cm} l(\chi(t_0;x,t_0,u(\cdot),w(\cdot)))
\end{align}
The value function $V(x,t)$ may be obtained by solving the HJ partial differential equation:
\begin{align}
\frac{\partial V}{\partial t} + H^{*} (x,\nabla V(x(t),t),t) = 0 \nonumber \\
V(x,0) = l(x), t\in[t_0,0] \nonumber \\
H^* = \underset{w(\cdot)}{\mbox{min}} \hspace{0.1cm} \underset{u(\cdot)} {\mbox{max}} \hspace{0.1cm} \nabla V(x,t)^T f(x,u)
\end{align}

\subsection{Control Lyapunov Function (CLF)}
The MPC control presented earlier assumes that there is no external disturbance. By design, MPC control can handle small disturbances but can fail in the presence of large disturbances. Here we present a Control Lyapunov Function to robustify the controller. 

Consider the nominal system obtained by setting $w(t) = 0$ in Eqn.~\ref{eq:actual_dynamic}
\begin{equation} \label{eq:nominal_dynamic}
\dot{\bar{x}}(t) = f(\bar{x}(t), \bar{u}(t), 0) 
\end{equation}

By defining the error \(e = x - \bar{x}\) between the nominal and actual system, the error dynamics can be described as:
\begin{equation} \label{eq:error_dynamic}
\dot{e} = f(x(t), u(t), w(t)) - f(\bar{x}(t), \bar{u}(t), 0)
\end{equation}

In order to achieve, $|e| \rightarrow 0$, we propose a robust controller of the form:
\begin{equation} \label{eq:proposed_controller}
u = \bar{u}_{\mbox{\tiny MPC}} + K(x-\bar{x})
\end{equation}
where \(\bar{u}_{\mbox{\tiny MPC}}\) is the feed-forward control from the MPC (see Sec.~\ref{sec:MPC}), $x$ is the measured state, $\bar{x}$ is the reference trajectory, and  \(  K(\cdot):=\mathbb{R}^n \to \mathbb{R}^m \) is the ancillary feedback control law derived from \textit{Lemma 2}. It can guarantee all solutions of (\ref{eq:error_dynamic}) are decreasing towards zero, and are inside an invariant set defined by \textit{Lemma 1}. 

\textbf{Lemma 1}: Suppose there exists \( E(x(t)) > 0 \), \( \lambda > 0 \) and \( \mu > 0 \) such that
\begin{equation} \label{eq:inv_inequa}
\frac{d}{dt} E(x(t)) + \lambda E(x(t))  -\mu^T w(t)w(t) < 0, 
\end{equation}
and if $\Omega(x)$ is the invariant set, then the system trajectory starting from \( x(t_0) \in \Omega(x) \) will remain in \( \Omega(x) \), where 
\begin{equation} \label{eq:INV}
\Omega(x) := \left\{ x \left| E(x(t)) < \frac{\mu w^2_{\text{max}}}{\lambda} \right. \right\}
\end{equation}

\textit{Proof}: See \citep{yu2010robust}. \hfill $\square$

\textit{Remark 1}: The size of the invariant set \(\Omega(x)\) may be tuned by adjusting \(\lambda\) and \(\mu\) to guarantee \(\Omega(x) \in \mathbb{X}\), that is, the invariant set is inside the feasible set of states.

\textit{Remark 2}: It is usually hard to model the uncertainty and find the disturbance bound \citep{cockburn1997linear}. In this paper, instead of directly modeling uncertainty, we utilize the Hamilton-Jacobi analysis (HJ) to determine the bound on the disturbance, \(\omega_{wax}\).

\textbf{Lemma 2}: Suppose there exist parameter scalars \( \lambda > 0 \) , \( \mu > 0 \) and matrix variables \( 0 < Y \in \mathbb{R}^{n \times n}, L \in \mathbb{R}^{m \times n} \), the cost function \(J = \int_{0}^{\infty} x^T Qx + u^T Ru \ \) can be minimized by the following optimization:
\begin{align} \label{eq:lemma2_0}
&\quad\max_{Y, L} \quad  \text{tr}(Y), \\
\text{s.t.} &\quad  \begin{bmatrix}
R_{11} & Y^T & L^T & B_w \\
Y & -Q^{-1} & 0 & 0 \\
L & 0 & -R^{-1} & 0 \\
B_w^T & 0 & 0 & -\mu I
\end{bmatrix} \prec 0, \\
&\quad R_{11}=(A Y + B L)^T + (A Y + B L) + \lambda Y,
\end{align}
where \(Q \in \mathbb{R}^{n \times n} \) and \(R \in  \mathbb{R}^{m \times m} \) are diagonal positive weight matrix. Then with \(u=Kx, E(x)=x^{T}Px\), where \(Y=P^{-1}\) and \(L=KY\), \textit{Lemma 1} is satisfied for the linearized form, \(\dot{x} = Ax + Bu + B_{w}w\), of the nonlinear system \(\dot{x}=f(x,u,w)\).

\textit{Proof}: Given in appendix.\hfill $\square$

\begin{figure}
\centering
\centering\includegraphics[scale=.9]{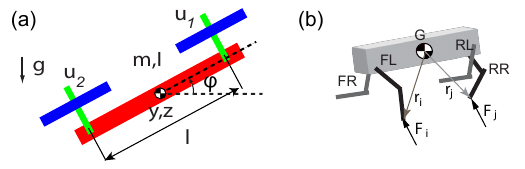}
\caption{(a) 2D quadcopter (b) 2D quadruped}
\label{fig:false-color}
\end{figure}

\subsection{Example 1: 2D Quadcopter}
\subsubsection{Nonlinear Model:}
The 2D Quadcopter is shown in Fig.~\ref{fig:false-color} (a) and the equations derived from using the Newton-Euler method are given below 
\begin{align} \label{eq:bicopter_dynamic}
 \dot{x} =  f(x,u,w) =
\begin{bmatrix}
\dot{y} \\
\dot{z} \\
\dot{\phi} \\
-m^{-1}u_{s}\sin{\phi} + w_1 \\
m^{-1}u_{s}\cos{\phi}-g + w_2\\
0.5I_{xx}^{-1} lu_{d}
\end{bmatrix}
\end{align}

where \(  y,z,\phi \) are horizontal, vertical position and orientation, \(\dot{y},\dot{z},\dot{\phi}\) are horizontal, vertical velocity and angular velocity, \(m, l,I_{xx}, g\) are mass, length, moment of inertia and gravity, \(u_{s} = u_1+u_2, u_{d} = u_1 - u_2 \) are the sum of the thrust forces and the difference in the thrust forces. We assume there is an additive disturbance \( w_i \in \mathbb{R} \), where $i=1,2$, and is bounded  \(\|w_i\|_{\infty} \leq w_{\scriptsize max} \).

\subsubsection{Linearized Model:}
The system may be linearized by assuming $\sin\phi \approx \phi$ and $\cos\phi \approx 1$, $u_s = mg + \delta u_s$, $u_d = \delta u_d$, and note that $\delta u_s sin\phi  = \delta u_s \phi \approx 0$
\begin{align} \label{eq:bicopter_linear_dynamic}
 \dot{x}  &= Ax + Bu + B_w w + G \nonumber \\
 \dot{\bar{x}}  &= A\bar{x} + B \bar{u} + G
\end{align}
where $A$ can be populated as follows (assuming index starts from 1): $A=zeros(6,6)$, $A(1,4)=A(2,5)=A(3,6) =1$, $A(4,3) = -g$; $B=zeros(6,2)$, $B(5,1)=1/m$ and $B(6,2) = 0.5I_{xx}^{-1} l$; $B_w = zeros(6,2)$, $B(4,1)=B(5,2)=1$, and $G=zeros(6,1)$, $G(5,1)=-g$, where the actual state is $x = \begin{bmatrix} y & z & \phi & \dot{y} & \dot{z} & \dot{\phi} \end{bmatrix}^T$, control is $u = \begin{bmatrix} \delta u_s & \delta u_d\end{bmatrix}^T$, the nominal state and control $\bar{x}$, $\bar{u}$ are similar to $x$, $u$ respectively, and the disturbance is $w = \begin{bmatrix} w_1 & w_2 \end{bmatrix}^T$. 

\subsubsection{Robust control:}
The proposed controller is of the form
\begin{equation} \label{eq:bicopter_controller}
u  = \bar{u}  + K(x - \bar{x})
\end{equation}
where the ancillary feedback control gain is \(K \in \mathbb{R}^{2 \times 6}  \) for the given $\lambda>0$ and $\mu>0$ and is computed from \textit{Lemma 2}. Furthermore, given the bound $w_{\mbox{\scriptsize max}}$, the region of attraction can be computed using \textit{Lemma 1}.

\begin{figure*}[tbp]
  \begin{center}
   \includegraphics[scale=1]{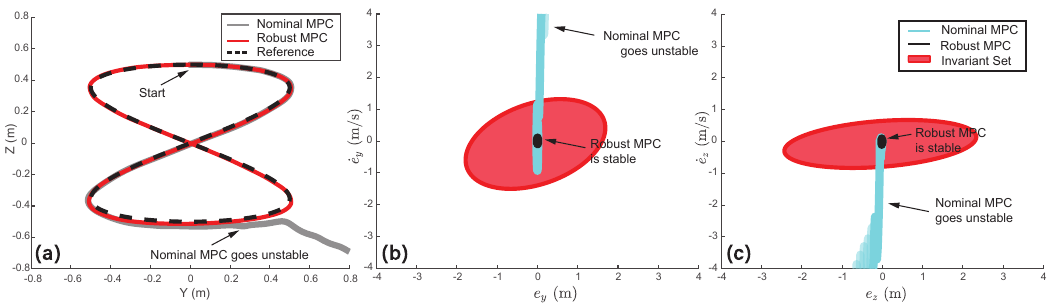}
 \end{center}
 \caption{Bicopter is tracking the figure 8 curve in the presence of bounded disturbance. (a) Figure 8 tracking in the y-z plane. (b) Horizontal tracking errors with invariant set. (c) Vertical tracking errors with invariant set.}
\label{fig:quadcopter-results}
\end{figure*}

\subsection{Example 2: 2D Quadruped}
\subsubsection{Nonlinear Model:}
The 2D quadruped is shown in Fig.~\ref{fig:false-color} (b) and the equations are derived from the Newton-Euler method. We have made the assumption that all mass is concentrated at the torso and the legs are massless. 
\begin{align} \label{eq:quadruped_dynamic}
 \dot{x} =  f(x,u,w) =
\begin{bmatrix}
\dot{y} \\
\dot{z} \\
\dot{\phi} \\
\bar{m}^{-1} ({\bf F}_{i} + {\bf F}_{j})- {\bf G} \\
 I_{xx}^{-1} ({\bf r}_{i} \times {\bf F}_{i} + {\bf r}_{j} \times {\bf F}_{j}) 
\end{bmatrix}
\end{align}

where the positions of the center of mass in the global frame are $y$ and $z$, the angular position of the body with respect to the horizontal direction is $\phi$, the ground reaction forces on the diagonal legs are ${\bf F}_i,{\bf F}_j \in \mathbb{R}^2$, where  $i$ and $j$ are diagonal legs, the position vectors from the respective feet to the center of mass are ${\bf r}_i,{\bf r}_j \in, \mathbb{R}^2$,  
the constant vector is $G=\{\mu g,g\}$ where gravity is $g$ and coefficient of static friction is $\mu$, inertia is $I_{xx} \in \mathbb{R}$, and the uncertainty is in the mass, $\bar{m} = (m + \Delta m) \in \mathbb{R}$, where the actual mass is $m$ and the uncertainty in mass is $\Delta m$. We assume that the quadruped moves in a trot gait where diagonal legs make contact with the ground. The step time is assumed to be constant. 

\subsubsection{Linearized Model:}
We assume that the linearized equation in the y- and z-direction (lines 1, 2, 4, 5 in Eqn.~\ref{eq:quadruped_dynamic}) can be written as 
\begin{align} \label{eq:quadruped_linear_dynamic}
 \dot{x}_i  &= A_ix_i + B_iu_i + B_{wi} w_i + G_i \nonumber \\
 \dot{\bar{x}}_i  &= A_i\bar{x}_i + B_i \bar{u} + G_i
\end{align}
where $x_i \in \mathbb{R}^2$, $i=1,2$ such that $x_1 = [y, \dot{y}]$ and $x_2 = [z, \dot{z}]$ $u_1 = \{\delta  Fx_{f}, \delta  Fx_{r}\}$ and $u_2 = \{\delta Fy_{f}, \delta Fy_{r}\}$ where $Fx, Fy$ are the control forces on the front and rear leg and subscript $f$ $r$ refer to the diagonal front and rear feet which are in contact with the groud, $A_i,B_i,B_{wi}$ can be populated as follows (assuming index starts from 1): $A_i=zeros(2,2)$, $A_i(1,2)=1$; $B=zeros(2,2)$, $B_i(2,1)=B_i(2,2) = 1/m$; $B_{wi} = zeros(2,1)$, $B_{wi}(2,1)=1$; $G_i = zeros(2,1)$, $G_1(2,1)=\mu, G_2(2,1)=- g$. We model the uncertainty in mass (see $\Delta m$ in Eqn~\ref{eq:quadruped_linear_dynamic}) as an additive disturbance, $w_i$, with a bound given by \(\mathbb{W} = \{w_i \in \mathbb{R} \mid \|w_i\|_{\infty} \leq w_{max}\}\). Here $w_{max}$ is unknown, but we will use HJ to estimate it as shown in HJ section below.

\subsubsection{Robust Control:}
The control is the same from \citep{bhounsule2023simple} but the stance phase control is modified for robustness as given below.
\begin{equation} \label{eq:quadruped_controller}
u_i = \bar{u}_i  + K(x_i - \bar{x}_i)
\end{equation}
where the ancillary feedback control gain is \(K \in \mathbb{R}^{2 \times 2}  \) for the given $\lambda>0$ and $\mu>0$ and is computed from \textit{Lemma 2}. Furthermore, given the bound $w_{\mbox{\scriptsize max}}$, the region of attraction may be computed using \textit{Lemma 1}. However, the bound $w_{\mbox{\scriptsize max}}$ is unknown, and we will use HJ to estimate it as shown in the HJ section below.

\subsubsection{Hamilton-Jacobi Reachability:}
The uncertainty in the model arises due to an unknown mass $\Delta m$ placed on top of the robot. Thus, the mass of the robot, \(\bar{m}=m  + \Delta m\), is uncertain. From Eqn~\ref{eq:quadruped_dynamic}, we observe that this is a multiplicative uncertainty that cannot be isolated symbolically as an additive term, $w$, as we have assumed in the linearized Eqn.~\ref{eq:quadruped_linear_dynamic}. We use HJ to identify a bound on $w$, $w_{\mbox{\scriptsize max}}$, as follows
\begin{enumerate}
\item The HJ uses the nonlinear dynamics Eqn.~\ref{eq:quadruped_dynamic} and the uncertainty (here the added mass $\Delta m$) to compute the Backward Reachable Set (BRS) \(\mathcal{R}\) for a given target set \(\mathcal{T}\). The safe set, $\mathcal{S}$, is the set of all states that are guaranteed to be invariant and are given by \(\mathcal{S}  =  \mathcal{R}  \cap \mathcal{T}   \). 
\item The CLF using {\textit Lemma 2} to compute the gain $K$ and then {\textit Lemma 1} to compute the region of attraction for a given $w_{\mbox{\scriptsize max}}$.
\item Now we combine HJ and CLF as follows. We compute the value of the worst-case uncertainty, $w_{\mbox{\scriptsize max}}$, using a line search method such that the region of attraction computed using the CLF lies within the safe set computed using HJ. 
\end{enumerate}

\begin{figure*}[tbp]
  \begin{center}
   \includegraphics[scale=0.95]{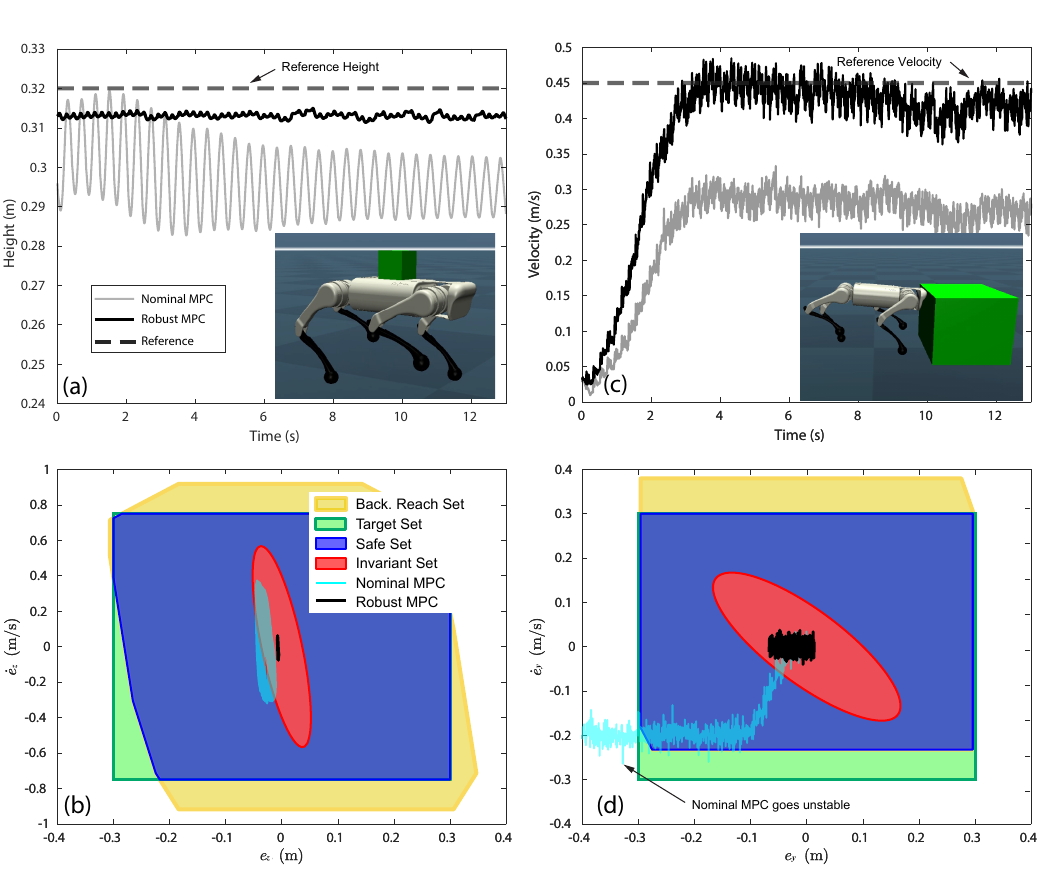}
 \end{center}
     \caption{For the 2D quadruped, two separate experiments are conducted. (a) simulation snapshot of quadruped while loading an unknown 5 kg object, along with the vertical height tracking result. (b) vertical height tracking errors with invariant set (c) simulation snapshot of quadruped while pushing an unknown 5 kg object along with the forward velocity tracking result (d) forward velocity tracking errors with invariant set}    
\label{fig:quadruped-results}
\end{figure*}

\section{Results and Discussion}

\subsubsection{Overview:}
We present results for the quadcopter tracking the figure 8 in the presence of disturbances and the quadruped following a reference velocity and maintaining a reference height in the presence of an added (unknown) mass and pushing a box. We compare the results of model predictive control and robust model predictive control that combines Control Lyapunov Function and Hamilton-Jacobi Reachability. 

For CLF, we used YALMIP \citep{lofberg2004yalmip}, a toolbox for solving Linear Matrix Inequality and other features in MATLAB, to compute the feedback gain ${\bf K}$ and positive definite matrix, ${\bf P}$. For HJ, we used \textit{helperOC} \citep{bansal2017hamilton}, optimal control toolbox for Hamilton-Jacobi Reachability Analysis, to compute the maximum disturbance $w_{\mbox{\scriptsize max}}$ for the quadruped. We used a custom simulation written in Python for 2D quadcopter and MuJoCo \citep{todorov2012mujoco} for the 2D quadruped. All computations are on on Ubuntu 20.04 on Intel Core i7.

\subsubsection{2D quadcopter:} The model parameters are $m=1$ kg, $l=0.2$ m, $I_{xx}=0.1$ kg$\cdot$m$^2$, $g=9.81$ m/s$^2$, and the additive disturbance $w_i \in \mathbb{W}$, where $\mathbb{W} = \{w_i \in \mathbb{R} \mid \|w_i\|_{\infty} \leq 3.5\}$ where $i=1,2$. For the quadcopter we track the figure 8 which is given by $x = 0.5\sin(2\tau)$ and $y=0.5\cos(\tau)$ where $\tau$ is assumed to be a fifth order polynomial and the constants are solved such that $\tau(t=0)=\dot{\tau}(t=0)=\ddot{\tau}(t=0)=\dot{\tau}(t=T)=\ddot{\tau}(t=T)=0$ and $\tau(t=T)=T$ where the end time is $T=5$. 

For the nominal MPC, the time step discretization is $dt = 0.05$ s, and the prediction horizon is two-time steps. The cost function matrices are ${\bf Q}_{MPC}=\mbox{diag}([100, 10, 1e9, 1e5, 1e14, 1e4])$ and ${\bf R}_{MPC} = \mbox{diag}([1e6, 1e6])$.

For the Robust MPC we used ${\bf R} = \mbox{diag}([1e-2, 1e-4])$, ${\bf Q} = \mbox{diag}([1e-1, 1, 1, 1, 1, 1e-2])$,  $\mu=0.1$, and $\lambda=0.5$. Using Lemma 1 and Lemma2 we computed the gain ${\bf K}$ and the positive definite matrix ${\bf P}$.

Figure~\ref{fig:quadcopter-results} (a) shows the quadcopter's ability to track the figure 8 using nominal MPC and robust MPC. It can be seen that the nominal MPC fails to track about halfway while the robust MPC is able to successfully track. Figure~\ref{fig:quadcopter-results} (b) and (c) shows the tracking error in the y- and z-direction.

\subsubsection{2D quadruped:} 
The model parameters $m = 12.454$ kg, $I_{xx} = 0.0565$ kg$\cdot m^2$, and $g = 9.81$ m/s$^2$, where the length of the thigh and shank are $\ell = 0.2$ m, consider an unknown mass disturbance $\Delta m = 5$ kg representing as the unknown mass loaded on the robot's torso and in front of the robot, interfering with height and velocity regulation. The nominal robot height is $0.32$ m and forward velocity is $0.45$ m/s.

For the nominal MPC, the time step discretization is $dt = 0.05$ s, and the MPC horizon is 2 time steps. The cost function matrices are ${\bf Q}_{MPC}=\mbox{diag}([1e5, 1e3, 1e7, 1e2, 1e1, 1e2])$ and ${\bf R}_{MPC} = {\bf 0}$.

For the control in the horizontal direction (y-direction) ${\bf R} = \mbox{diag}([1])$, ${\bf Q} = \mbox{diag}([500,10])$,  $\mu=200$, and $\lambda=0.3$. Using Lemma 1 and Lemma 2 we computed the gain ${\bf K}$ and the positive definite matrix ${\bf P}$
For the control in the vertical direction (z-direction), ${\bf R} = \mbox{diag}([0.01])$, ${\bf Q} = \mbox{diag}([1000,1])$,  $\mu=90$, and $\lambda=0.8$. Using Lemma 1 and Lemma 2 we computed the gain ${\bf K}$ and the positive definite matrix ${\bf P}$

Figure~\ref{fig:quadruped-results} (a) and (b) shows reference height tracking in a trot gait with added weight. The reference height is the dashed gray line. The robust MPC (black line) shows substantially lower tracking error than nominal MPC (gray line). Figure~\ref{fig:quadruped-results} (c) and (d) show reference velocity tracking when the robot is pushing a weight. The reference velocity is shown as the dashed gray line. The robust MPC (black line) shows substantially low tracking error compared to nominal MPC (gray line). The latter eventually goes beyond the safe region while the robust MPC stays within the Invariant set.

In this paper, we have shown that CLF can be combined with HJ reachability to create a robust controller with safety guarantees for an existing nominal controller (here MPC). The method capitalizes on convex optimization with a quadratic cost and a linear model to compute the RoAs as a function of the additive disturbances. Then HJ is used to calculate the correct approximation for the additive disturbance to select the correct ROA. The results are demonstrated on a 2D quadcopter and 2D quadruped in simulation. 

The method has its limitations. The process is conservative and requires extensive trial and error to compute the gains and invariant set for the CLF. The technique is strictly offline and relies on a good understanding of the nonlinear model and the model uncertainty. The technique works well for fully actuated or under-actuated systems without strong coupling (e.g., for 2D quadruped we assumed the coupling between pitching motion and translation was minimal).


\bibliography{ifacconf}             

\appendix
\section{Proof of Lemma 2}    

We prove that \(\textit{Lemma 2}\) holds for the system as follow:
\begin{equation} \label{eq:subsystem1}
\dot{x} = Ax + Bu + B_{\omega}\omega
\end{equation}
\begin{equation} \label{eq:subsystem_control}
u = Kx
\end{equation}

Firstly, to achieve equilibrium point tracking, an optimization problem is formulated. This involves using a standard infinite horizon quadratic cost as the performance measure.
\begin{equation} \label{eq:LQR_cost}
\min \int_{0}^{\infty}x^TQx + u^TRu
\end{equation}

A controller which minimizes the proposed cost is given by solving the following semidefinite programming problem:
\begin{equation} \label{eq:proof1}
\mathcal{P} : \min_{P,K} \quad \text{tr}(P)
\end{equation}
\begin{equation} \label{eq:proof2}
s.t. \quad (A + BK)^TP + P(A + BK) \prec -Q - K^TRK
\end{equation}

where positive definite matrix \(P\) is defined by  the propose CLF \(E(x)=x^TPx\)  and  \(Q, R\) are diagonal weight matrices. Assuming the existence of positive scalars \(\mu\) and  \(\lambda\) , the original constraint can be rewritten to include additional negative terms \( -PB_{\omega}(\mu)^{-1}(PB_{\omega})^T \) , \(-\lambda P\) . This modification preserves the validity of  \((\ref{eq:proof2})\) the original constraint, leading to the new formulation as follows:
\begin{multline} \label{eq:schur}
\quad (A + BK)^TP + P(A + BK) \prec \\
-Q - K^TRK -PB_{\omega}(\mu)^{-1}(PB_{\omega})^T  - \lambda P   
\end{multline}

Using Schur complement,  the constraint (\ref{eq:schur}) can be rewritten as:
\begin{equation} \label{eq:schur1}
\bullet := (A+BK)^TP + P(A+BK)+\lambda P
\end{equation}
\begin{equation} \label{eq:schur2}
\begin{bmatrix}
 \bullet&  I&  K^T&  PB_{\omega} \\
 I&  (-Q)^{-1}&  0&  0\\
 K&  0&  (-R_{1})^{-1}& 0 \\
 B_{\omega}^TP&  0&  0&  -\mu\\

\end{bmatrix} \prec 0
\end{equation}

Performing a congruence transformation on (\ref{eq:schur2}) with the matrix \(diag(x, \Xi, \Phi, \omega)\), the following inequality is obtained: 
\begin{multline} \label{eq:schur3}
x^T[(A+BK)^TP + P(A+BK)  +  \lambda P ]x + \\
\Xi^TIx + \Phi^TKx +  \omega^T(PB_{\omega})^Tx + 
x^TI\Xi - \Xi^{T} Q^{-1}\Xi + \\
x^TK^T\Phi - \Phi^T R^{-1} \Phi + 
x^TPB_{\omega}\omega - \mu\omega^T\omega < 0
\end{multline}

Assuming \(\Xi, \Phi = 0\), and considering the CLF \(E(x)=x^TPx\) along with its derivative  \(\frac{d}{dt}E(x)=\dot{x}^TPx + x^TP\dot{x}\) , (\ref{eq:schur3}) can be reformulated as follows:
\begin{equation} \label{eq:proof_inv}
\frac{d}{dt}E(x) + \lambda E(x)-\mu \omega^T\omega < 0
\end{equation}

Thus, according to \textit{Lemma 1}, by solving semidefinite programming problem \(\mathcal{P}_{1}\) subject to constraint (\ref{eq:proof2}), it can be guaranteed that system remains robustly invariant within the set  \( \Omega(x) := \left\{ x \left| E(x) < \frac{\mu w^2_{\text{max}}}{\lambda} \right. \right\} \) .

By using standard method in control, performing a congruence transformation with \(Y=P^{-1}\), introducing \(L=KY\), and applying a Schur complement, nonconvex problem \(\mathcal{P}\) can be formulated as a convex problem as follow:
\begin{equation} \label{eq:lemma2_proof1}
\max_{Y, L} \quad  \text{tr}(Y) 
\end{equation}
\begin{equation} \label{eq:lemma2_proof2}
*:=(A Y + B L)^T + (A Y + B L) + \lambda Y
\end{equation}
\begin{equation} \label{eq:lemma2_proof3}
\text{s.t.} \quad  \begin{bmatrix}
* & Y^T & L^T & B_{\omega} \\
Y & -Q^{-1} & 0 & 0 \\
L & 0 & -R^{-1} & 0 \\
B_{\omega}^T & 0 & 0 & -\mu
\end{bmatrix} \prec 0
\end{equation}



\end{document}